\documentclass[conference]{IEEEtran}
\IEEEoverridecommandlockouts
\usepackage{cite}
\usepackage{amsmath,amssymb,amsfonts,upgreek}
\usepackage{algorithmic}
\usepackage{graphicx}
\usepackage{textcomp}
\usepackage{xcolor}
\usepackage{url}
\usepackage{multirow}
\usepackage[caption=false,font=normalsize,labelfont=sf,textfont=sf]{subfig}
\def\BibTeX{{\rm B\kern-.05em{\sc i\kern-.025em b}\kern-.08em
    T\kern-.1667em\lower.7ex\hbox{E}\kern-.125emX}}
\begin{document}

\title{Towards End-to-End GPS Localization with Neural Pseudorange Correction
\thanks{This work was supported by NTU Research Scholarship.}
}

\author{
\IEEEauthorblockN{Xu Weng}
\IEEEauthorblockA{\textit{School of Electrical and Electronic Engineering} \\
\textit{Nanyang Technological University}\\
Singapore, Singapore \\
xu009@e.ntu.edu.sg}
\and
\IEEEauthorblockN{K.V. Ling}
\IEEEauthorblockA{\textit{School of Electrical and Electronic Engineering} \\
\textit{Nanyang Technological University}\\
Singapore, Singapore \\
ekvling@ntu.edu.sg}
\and
\IEEEauthorblockN{Haochen Liu}
\IEEEauthorblockA{\textit{School of Mechanical and Aerospace Engineering} \\
\textit{Nanyang Technological University}\\
Singapore, Singapore \\
haochen002@e.ntu.edu.sg}
\and
\IEEEauthorblockN{Kun Cao}
\IEEEauthorblockA{\textit{School of Electrical Engineering and Computer Science} \\
\textit{KTH Royal Institute of Technology}\\
Stockholm, Sweden \\
kun001@e.ntu.edu.sg; caokun@kth.se}
}

\maketitle

\begin{abstract}
The pseudorange error is one of the root causes of localization inaccuracy in GPS. Previous data-driven methods regress and eliminate pseudorange errors using handcrafted intermediate labels. Unlike them, we propose an end-to-end GPS localization framework, E2E-PrNet, to train a neural network for pseudorange correction (PrNet) directly using the final task loss calculated with the ground truth of GPS receiver states. The gradients of the loss with respect to learnable parameters are backpropagated through a Differentiable Nonlinear Least Squares (DNLS) optimizer to PrNet. The feasibility of fusing the data-driven neural network and the model-based DNLS module is verified with GPS data collected by Android phones, showing that E2E-PrNet outperforms the baseline weighted least squares method and the state-of-the-art end-to-end data-driven approach. Finally, we discuss the explainability of E2E-PrNet.
\end{abstract}


\section{Introduction}
Pseudorange errors are a long-standing curse of GPS localization, resulting in positioning errors that are hard to mitigate. Because of their complicated composition, encompassing satellite clock errors, atmospheric delays, receiver noise, hardware delays, and so on \cite{kaplan2017understanding}, how to remove them has long been under active research in the GPS community. Numerous mathematical and experimental models have been developed to remove various components in pseudorange errors \cite{kaplan2017understanding}. However, the community is still bothered about some tough stains in pseudorange measurements, such as multipath/non-line of sight (NLOS) errors, modeling residual errors, and user hardware biases. These problems are particularly severe in low-cost GPS receivers, such as those mounted in mass-market smartphones. Due to the difficulty of modeling such pseudorange errors mathematically, researchers have been falling back on big data to attend to this issue.

On the one hand, most previous work performed supervised learning to regress pseudorange errors using handcrafted labels--we only have the ground truth of locations of GPS receivers and have to derive the target values for pseudorange errors using our domain knowledge of GPS. Various derived labels of pseudorange errors are proposed, including pseudorange errors containing residual receiver clock offset \cite{sun2020improving}, double difference of pseudoranges \cite{zhang2021prediction}, and smoothed pseudorange errors \cite{weng2024prnet}. However, the final task target values--receiver locations--are in place but not used directly. On the other, end-to-end deep learning approaches are proposed to directly map GPS measurements to user locations and implicitly correct pseudorange errors \cite{kanhere2022improving,mohanty2022learning,xu2023positionnet}. However, these approaches put considerable but unnecessary commitment to learning well-established and robust classical localization theories.
\begin{figure*}[!t]
    \centering
    \includegraphics[width=0.9\textwidth]{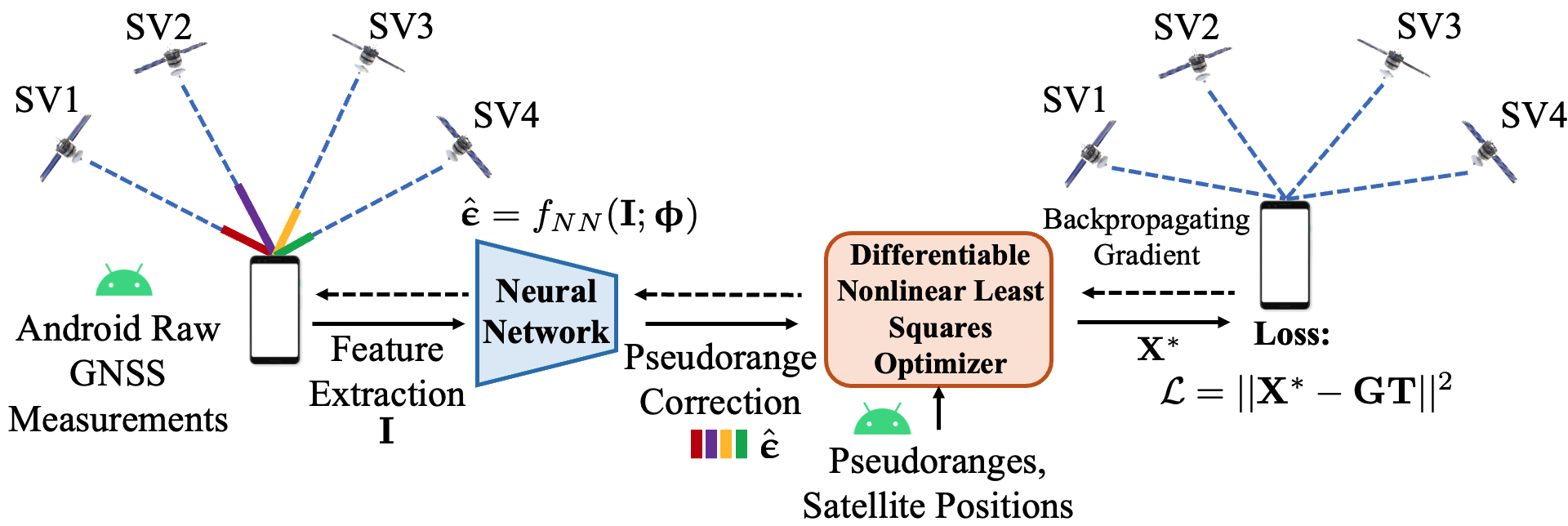}
    \caption{An overview of our E2E-PrNet GPS localization pipeline. The learnable parameters $\boldsymbol{\upphi}$ of the neural network for pseudorange correction are tuned by the final task loss calculated using the receiver states $\mathbf{X}^*$.}
    \label{fig: overview}
\end{figure*}

The data-driven methods can learn pseudorange errors while the model-based approaches can perfectly compute locations using the corrected pseudoranges. Thus, can we \textbf{fuse} them so that we can preserve our well-established priors and train the neural modules using the final task loss instead of the intermediate loss? Such hybrid end-to-end pipelines have achieved success in many other domains, including synthesis of novel views \cite{mildenhall2021nerf}, object pose estimation \cite{chen2022epro}, robot control \cite{wang2021differentiable}, autonomous driving \cite{huang2023differentiable}, and so on. In the GPS community, recently, an end-to-end learning framework has been proposed to improve GPS positioning using the differentiable factor graph optimizer, the factor weightings of which are tweaked by the final task loss of receiver locations \cite{xu2023differentiable}.\IEEEpubidadjcol

This paper proposes E2E-PrNet, an end-to-end GPS localization framework with learnable pseudorange correction. Our main contributions are listed below.
\begin{itemize}
\item As shown in Fig. \ref{fig: overview}, we use a neural network (PrNet) to regress pseudorange errors that we then combine with other necessary measurements and feed into a Differentiable Nonlinear Least Squares (DNLS) optimizer for location computation. We calculate the loss using the state estimates of the DNLS optimizer and the ground truth of receiver states, the gradients of which are backpropagated through the DNLS optimizer to tune the learnable parameters of PrNet. 

\item To handle the issue of having no target values for receiver clock offsets, we select their Weighted Least Squares (WLS)-based estimates to label them. 

\item We evaluate our proposed pipeline using Google Smartphone Decimeter Challenge (GSDC) datasets and compare it with the baseline WLS algorithm and a state-of-the-art (SOTA) end-to-end approach. In this work, we only focus on GPS data, but our framework can readily be extended to other constellations.

\item Finally, we explore what the front-end PrNet has learned when trained by the final task loss. The codes of E2E-PrNet are available at \url{https://github.com/ailocar/e2eprnet}.
\end{itemize}

To the best of our knowledge, our proposed E2E-PrNet is the first data-driven GPS pseudorange correction pipeline trained with the final task loss in an end-to-end way. By fusing the data-driven and model-based modules, E2E-PrNet obtains superior localization performance to its competitors.  

\section{Preliminaries of GPS}
To estimate the unknown location $\mathbf{x}_k=[x_k, y_k, z_k]^T$ and the clock offset $\delta t_{u_k}$ of a GPS receiver at the $k^{th}$ epoch, we need to solve a system of $M$ nonlinear pseudorange equations that measure the distances from the receiver to $M$ visible satellites. The pseudorange of the $n^{th}$ satellite at the $k^{th}$ epoch can be modeled as $\rho_{k}^{(n)}$:
\begin{equation}\label{eq:PrEq}
\rho_{k}^{(n)}=||\mathbf{x}_k-\mathbf{x}_k^{(n)}||+\delta t_{u_k}+\varepsilon_{k}^{(n)}
\end{equation}
where $\mathbf{x}_k^{(n)}$ denotes the satellite position. In \eqref{eq:PrEq}, we have modeled and removed the satellite clock error, the relativistic effect correction, and the group delay using the information stored in navigation messages. The ionospheric delay is mitigated with the Klobuchar model. The tropospheric delay $T_k^{(n)}$ is modeled as
\begin{equation}
T_k^{(n)}=\frac{2.47}{0.0121+\sin E_k^{(n)}}\nonumber
\end{equation}
where $E_k^{(n)}$ is the elevation angle of the satellite at the receiver \cite{kaplan2017understanding}. Therefore, the measurement error $\varepsilon_{k}^{(n)}$ only includes the multipath/NLOS delays, modeling residual errors, hardware delays, pseudorange noise, etc. The measurement error can be considered as the sum of the denoised pseudorange error $\mu_{k}^{(n)}$ and the unbiased pseudorange noise $\upsilon_{k}^{(n)}$, i.e., $\varepsilon_{k}^{(n)}=\mu_{k}^{(n)}+\upsilon_k^{(n)}$. With an approximation to the receiver's state $\tilde{\mathbf{X}}_k=[\tilde{x}_k, \tilde{y}_k, \tilde{z}_k, \delta \tilde{t}_{u_k}]^T$, we can apply the Gauss-Newton algorithm to calculate the least squares (LS) estimation of the receiver's state $\hat{\mathbf{X}}_k=[\hat{x}_k, \hat{y}_k, \hat{z}_k, \delta \hat{t}_{u_k}]^T$:
\begin{equation}\label{eq:GNeq}
    \hat{\mathbf{X}}_k=\tilde{\mathbf{X}}_{k}-\left(\mathbf{J}_{\mathbf{r}_k}^T\mathbf{J}_{\mathbf{r}_k}\right)^{-1}\mathbf{J}_{\mathbf{r}_k}^T\mathbf{r}_k(\tilde{\mathbf{X}}_{k})
\end{equation}
where 
\begin{gather}
\mathbf{r}_k(\tilde{\mathbf{X}}_{k})=\left[\tilde{r}_k^{(1)},\tilde{r}_k^{(2)},...,\tilde{r}_k^{(M)}\right]^T \nonumber
\\
\tilde{r}_k^{(n)}=\rho_{k}^{(n)}-||\tilde{\mathbf{x}}_k-\mathbf{x}_k^{(n)}||-\delta \tilde{t}_{u_k} \nonumber
\\
\mathbf{J}_{\mathbf{r}_k}=\left[\frac{\partial{\mathbf{r}_k}(\tilde{\mathbf{X}}_{k})}{\partial \mathbf{\tilde{X}}_k}\right]_{M\times 4}\nonumber
\end{gather}

We run \eqref{eq:GNeq} iteratively until a certain criterion is fulfilled. Let $\mathbf{H}_k = \left(\mathbf{J}_{\mathbf{r}_k}^T\mathbf{J}_{\mathbf{r}_k}\right)^{-1}\mathbf{J}_{\mathbf{r}_k}^T$ and ${\mathbf{X}}_k$ represent the true receiver state. We assume $\mathbf{H}_k$ has been weighted implicitly using pseudorange uncertainty. Then, the WLS estimation error $\mathbf{\epsilon}_{\mathbf{X}_k}$ is \cite{weng2024prnet}:
\begin{equation}\label{eq:exk} \mathbf{\epsilon}_{\mathbf{X}_k}=\mathbf{X}_k-\hat{\mathbf{X}}_k=-\mathbf{H}_k\boldsymbol{\varepsilon}_k  
\end{equation}
where $\boldsymbol{\varepsilon}_k=[\varepsilon_{k}^{(1)},\varepsilon_{k}^{(2)},\cdots,\varepsilon_{k}^{(M)}]^T$.

\begin{figure}[t]
\centerline{\includegraphics[width=0.5\textwidth]{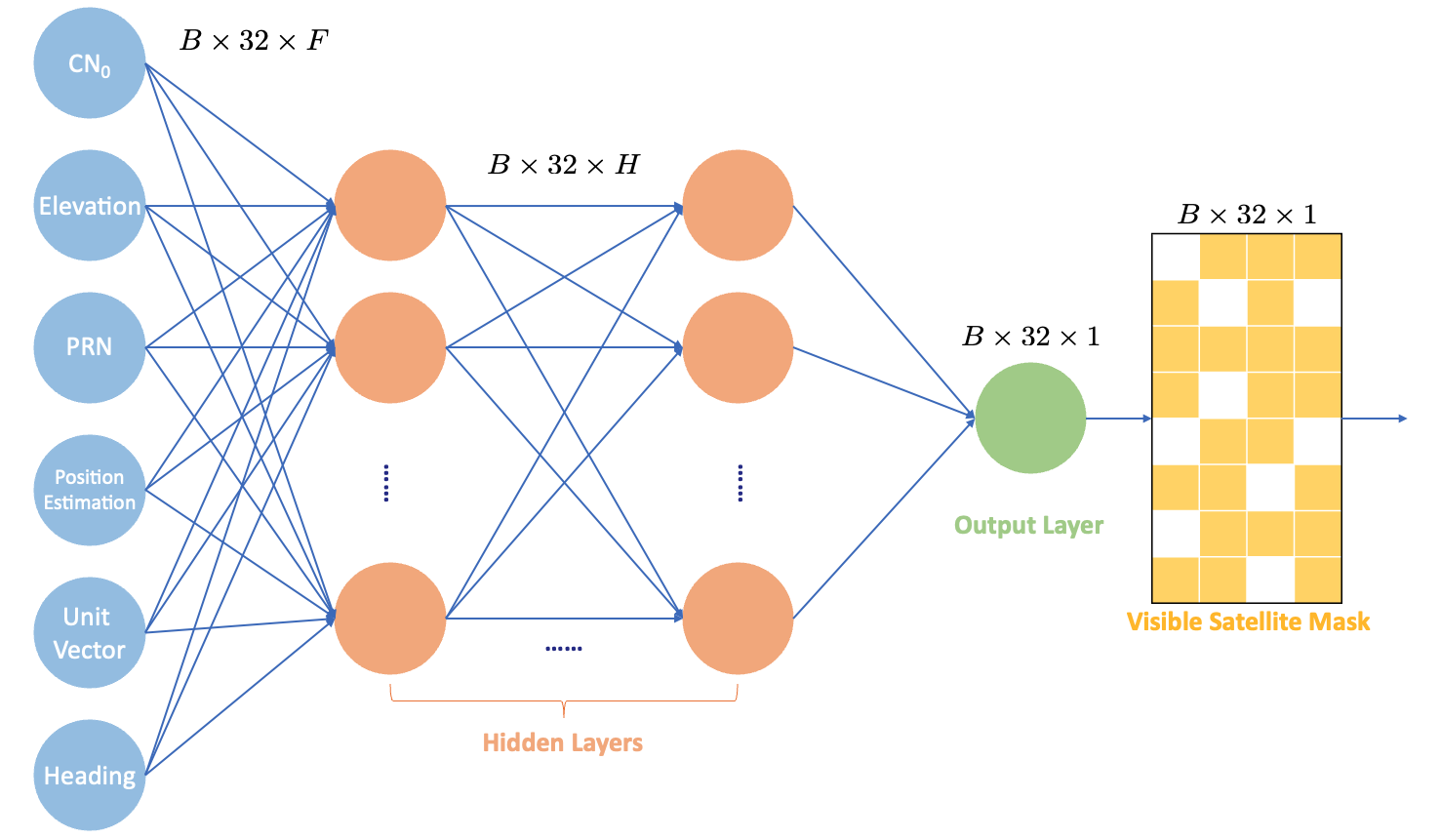}}
\caption{The diagram of PrNet. PrNet consists of a MLP layer for pseudorange error regression and a mask layer for filtering out visible satellites. $B$ represents the batch size. $F$ denotes the feature dimension. $H$ is the number of hidden neurons.}
\label{fig:prNet}
\end{figure}

\begin{figure*}[t]
\centerline{\includegraphics[width=1\textwidth]{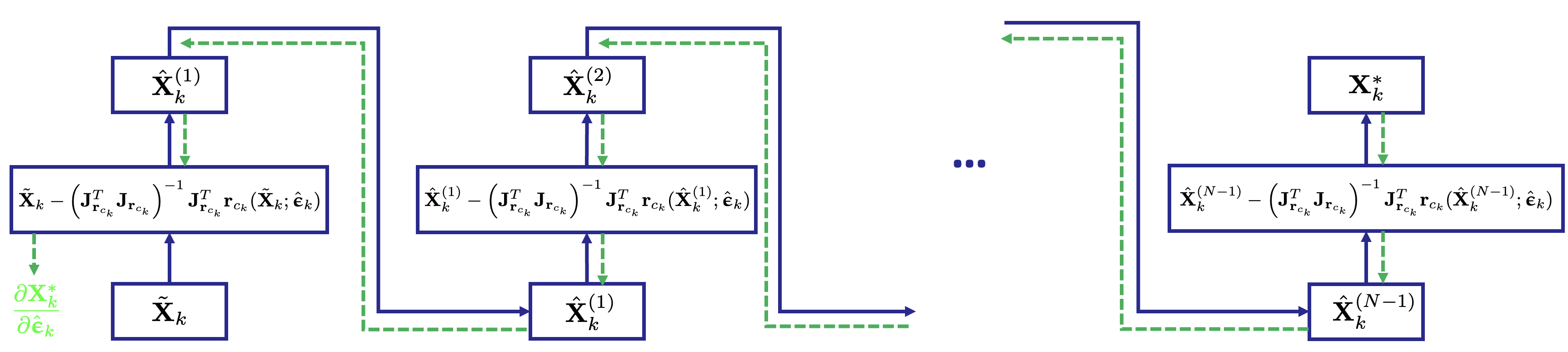}}
\caption{The computational graph of the unrolling Gauss-Newton iteration: the forward process (in blue lines) and the backward process (in green dash lines) }
\label{fig:comGraph}
\end{figure*}

\section{Methodology}
As \eqref{eq:exk} shows, a key to improving GPS localization accuracy is to reduce the pseudorange measurement errors. To this end, we propose an end-to-end learning pipeline (E2E-PrNet) to train a neural network for pseudorange correction using the final task loss of GPS. As shown in Fig. \ref{fig: overview}, we connect the output of a neural network to a DNLS optimizer for state estimation. The loss is calculated with the estimated and true receiver states, and its gradients are backpropagated through the DNLS optimizer to the neural network to tune its learnable parameters. 

\subsection{Neural Pseudorange Correction}
We employ PrNet as the front-end neural network for correcting pseudoranges considering its SOTA performance \cite{weng2024prnet}. As shown in Fig. \ref{fig:prNet}, PrNet is composed of a basic Multilayer Perceptron (MLP) followed by a mask layer to regress the pseudorange errors from six satellite, receiver, and context-related input features:
\begin{equation*}
    \hat{\varepsilon}_k^{(n)}=f_{NN}(\mathbf{I}_k^{(n)};\mathbf{\Phi})
\end{equation*}
where $\mathbf{I}_k^{(n)}$ represents the input features for the $n^{th}$ satellite, including the carrier-to-noise density ratio ($C/N_0$), its elevation angle, its PRN index, the WLS-based receiver position estimate, the unit geometry vector from it to the receiver, and the receiver heading estimate. $\mathbf{\Phi}$ is the vector of learnable parameters of the neural network. In the original framework of PrNet, target values of pseudorange errors are manually derived from the receiver location ground truth as follows \cite{weng2024prnet}.
\begin{equation}\label{eq:Prlabel}
    \mu_k^{(n)}-\mathbf{h}_{t_k}^T\mathbf{M}_k=||\bar{\mathbf{x}}_k-\mathbf{x}_k^{(n)}||-||{\mathbf{x}}_k-\mathbf{x}_k^{(n)}||
\end{equation}
where $\mathbf{h}_{t_k}^T$ is the last row vector of the matrix $\mathbf{H}_k$. The vector $\mathbf{M}_k$ represents the denoised pseudorange errors of all visible satellites, i.e., $\mathbf{M}_k=[\mu_k^{(1)},\mu_k^{(2)},\cdots,\mu_k^{(M)}]^T$. $\bar{\mathbf{x}}_k$ is the smoothed estimation of the receiver location. Note that the common delay item $-\mathbf{h}_{t_k}^T\mathbf{M}_k$ for all visible satellites does not affect the localization accuracy.

\subsection{Differentiable Nonlinear Least Squares Optimizer}
We consider GPS localization as a nonlinear least squares optimization problem:
\begin{equation}\label{eq:DiffLoc}
\mathbf{X}_k^*=\min\limits_{\hat{\mathbf{X}}_k}\frac{1}{2}||\mathbf{r}_{c_k}(\hat{\mathbf{X}}_k;\hat{\boldsymbol{\upepsilon}}_k)||^2
\end{equation}
where 
\begin{gather}
\mathbf{r}_{c_k}(\hat{\mathbf{X}}_k;\hat{\boldsymbol{\upepsilon}}_k)=\left[r_{c_k}^{(1)},r_{c_k}^{(2)},...,r_{c_k}^{(M)}\right]^T \nonumber
\\
r_{c_k}^{(n)}=\rho_{k}^{(n)}-\hat{\varepsilon}_k^{(n)}-||\hat{\mathbf{x}}_k-\mathbf{x}_k^{(n)}||-\delta \hat{t}_{u_k}. \label{eq:residual}   
\end{gather}
The optimization variables include the receiver's location and clock offset, i.e., $\hat{\mathbf{X}}_k=[\hat{x}_k, \hat{y}_k, \hat{z}_k, \delta \hat{t}_{u_k}]^T$. The auxiliary variables are satellite locations $[\mathbf{x}_k^{(n)}]_M$, pseudorange measurements $[\rho_k^{(n)}]_M$, and neural pseudorange corrections $\hat{\boldsymbol{\upepsilon}}_k=[\hat{\varepsilon}_k^{(n)}]_M$. The data-driven and model-based modules are connected via \eqref{eq:residual}. Then, the final task loss is computed with the optimized and true receiver state $\mathbf{X}_k$:
\begin{equation*}
    \mathcal{L}=||\mathbf{X}_k^*-\mathbf{X}_k||^2.
\end{equation*}

The derivative of the loss function with respect to the learnable parameters $\mathbf{\Phi}$ of the front-end neural network is calculated as
\begin{equation} \label{eq:lossDiff}
    \frac{\partial \mathcal{L}}{\partial \mathbf{\Phi}}=\frac{\partial \mathcal{L}}{\partial \mathbf{X}_k^*}\cdot\frac{\partial \mathbf{X}_k^*}{\partial \hat{\boldsymbol{\upepsilon}}_k}\cdot\frac{\partial \hat{\boldsymbol{\upepsilon}}_k}{\partial \mathbf{\Phi}}
\end{equation}
where the first derivative of the right side of \eqref{eq:lossDiff} is easy to compute considering its explicit form. The last one can be solved in a standard training process. But computing the derivative of the optimal state estimation $\mathbf{X}_k^*$ with respect to the neural pseudorange correction $\hat{\boldsymbol{\upepsilon}}_k$ is challenging due to the differentiation through the nonlinear least squares problem. 

To solve \eqref{eq:DiffLoc}, we can substitute $r_k(\tilde{\mathbf{X}}_k)=r_{c_k}(\tilde{\mathbf{X}}_k;\hat{\boldsymbol{\upepsilon}}_k)$ into \eqref{eq:GNeq} and perform the Gauss-Newton algorithm iteratively for $N$ times such that the state estimate converges, which is illustrated by the unrolling computational graph in blue lines in Fig. \ref{fig:comGraph}. Accordingly, the gradients of $\mathbf{X}_k^*$ with respect to $\hat{\boldsymbol{\upepsilon}}_k$ can be calculated in the backward direction along the computational graph, as shown by the green dash lines in Fig. \ref{fig:comGraph}. This is the basic idea behind differentiable nonlinear least squares optimization for GPS localization. We use Theseus, a generic DNLS library, to solve the optimization problem \eqref{eq:DiffLoc} and calculate the derivative \eqref{eq:lossDiff} since it can backpropagate gradients like \eqref{eq:lossDiff} using various algorithms, such as the unrolling differentiation displayed by Fig. \ref{fig:comGraph} \cite{pineda2022theseus}.  

\subsection{Handling the Missing Label Issue}
In practice, we can collect the ground truth of user locations using high-performance geodetic GPS receivers integrated with other sensors, such as visual-inertial localization systems. However, the ground truth of receiver clock offsets is difficult to obtain. Not supervising receiver clock offsets may lead to arbitrary shared biased errors in neural pseudorange corrections (the receiver clock offset can {\it absorb} shared pseudorange errors among visible satellites \cite{kaplan2017understanding}), leaving the neural network less interpretable. To deal with the issue, we choose the WLS-based estimation of the receiver clock offset $\delta \hat{t}_{u_k}$ as the target value of $\delta {t}_{u_k}$:
\begin{equation}\label{eq:lsDtu}
    \delta \hat{t}_{u_k}=\delta {t}_{u_k}+\mathbf{h}_{t_k}^T\boldsymbol{\varepsilon}_k
\end{equation}

Considering a perfectly trained E2E-PrNet, i.e., its output is exactly the same as the ground truth, we can get the following equations according to \eqref{eq:PrEq}, \eqref{eq:DiffLoc}, \eqref{eq:residual}, and \eqref{eq:lsDtu}.
\begin{IEEEeqnarray}{rCl}\label{eq:labelE2EPrNet}
\hat{\varepsilon}_k^{(n)}&=&{\varepsilon}_k^{(n)}-\mathbf{h}_{t_k}^T\boldsymbol{\varepsilon}_k \nonumber
\\
&=&\mu_k^{(n)}-\mathbf{h}_{t_k}^T\mathbf{M}_k+\upsilon_k^{(n)}-\mathbf{h}_{t_k}^T\mathbf{\Upsilon}_k
\end{IEEEeqnarray}
where $\mathbf{\Upsilon}_k=[\upsilon_k^{(1)},\upsilon_k^{(2)},\cdots,\upsilon_k^{(M)}]^T$ is the unbiased pseudorange noise of all visible satellites. Therefore, by comparing \eqref{eq:Prlabel} and \eqref{eq:labelE2EPrNet}, we can conclude that the proposed E2E-PrNet is equivalent to a PrNet trained with noisy pseudorange errors.

\begin{figure}[!t]
\centering
\subfloat[]{\includegraphics[width=0.24\textwidth]{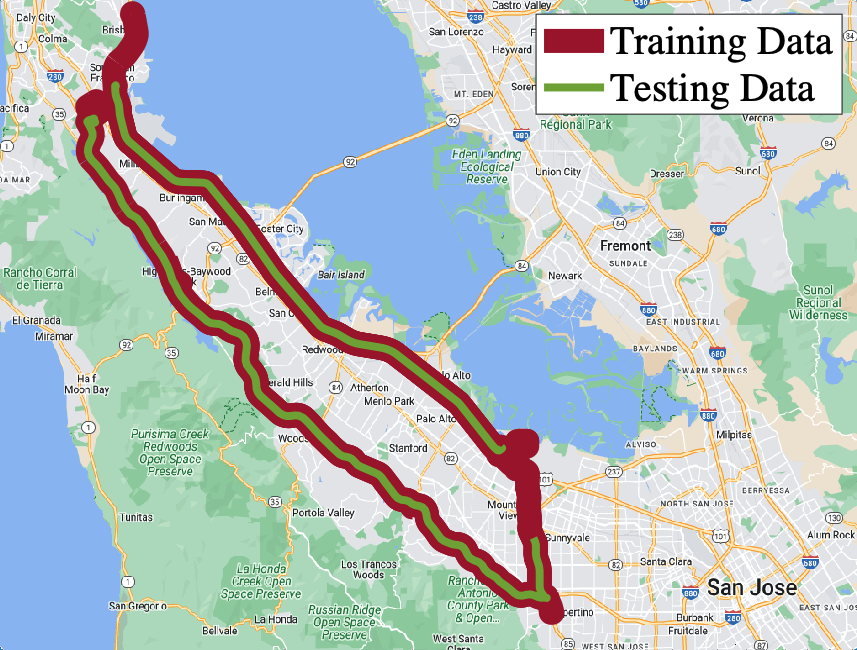}\label{fig: Scenario1}}
\hfil
\subfloat[]{\includegraphics[width=0.24\textwidth]{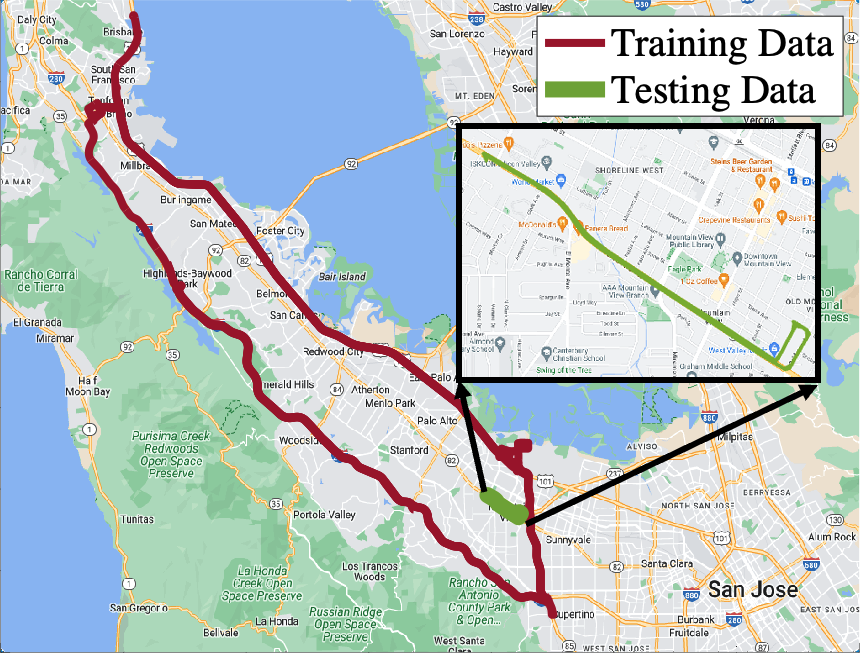}\label{fig: Scenario2}}
\caption{(a) Scenario I: fingerprinting. (b) Scenario II: cross trace.}
\label{fig:Scenarios}
\end{figure}

\begin{figure}[t]
\centering
\subfloat[]{\includegraphics[width=0.48\textwidth]{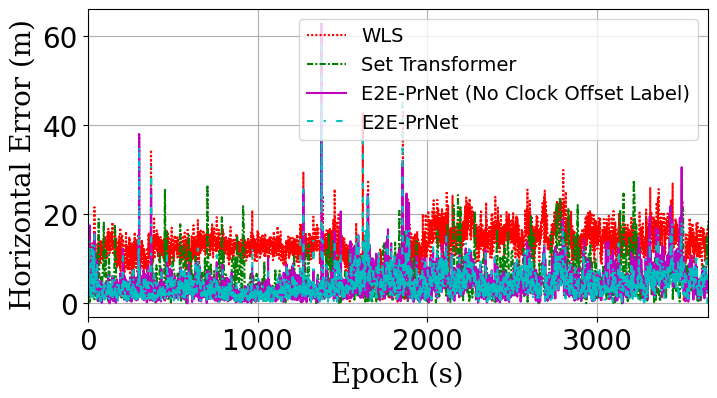}\label{fig: RMSE1}}
\hfil
\subfloat[]{\includegraphics[width=0.48\textwidth]{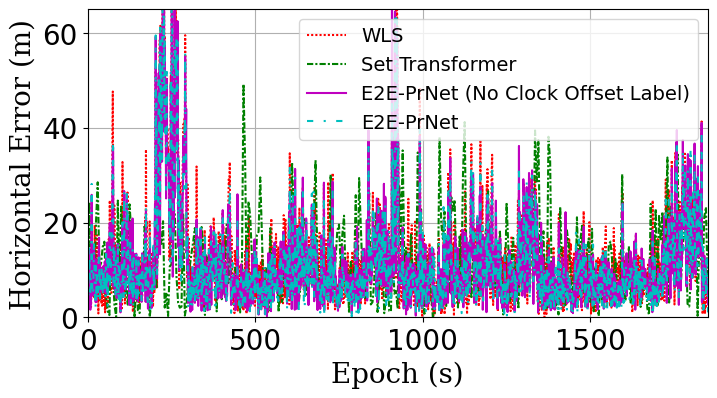}\label{fig: RMSE2}}
\caption{Horizontal errors in (a) Scenario I. (b) Scenario II.}
\label{fig:RMSE}
\end{figure}

\begin{figure}[t]
\centering
\subfloat[]{\includegraphics[width=0.48\textwidth]{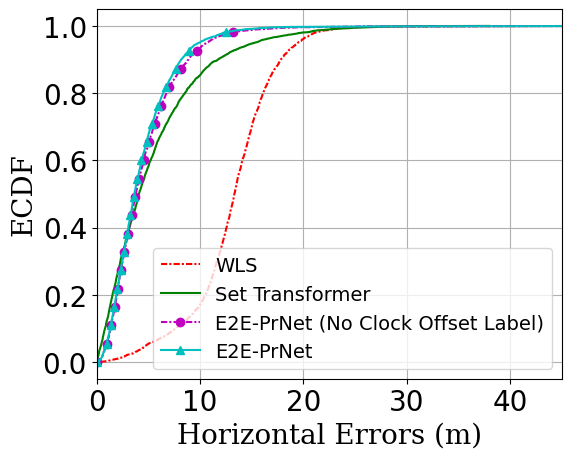}\label{fig: ECDF1}}
\hfil
\subfloat[]{\includegraphics[width=0.48\textwidth]{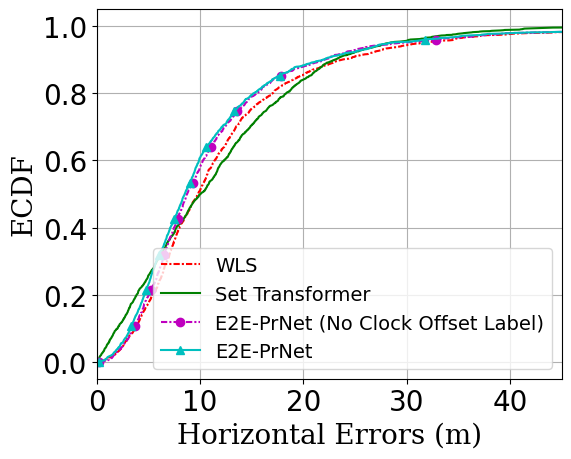}\label{fig: ECDF2}}
\caption{ECDF of horizontal errors in (a) Scenario I. (b) Scenario II.}
\label{fig:ECDF}
\end{figure}

\begin{table}[!t]
\caption{Details of Data Sets}
\centering
\begin{tabular}{c|c|c|c|c}
\hline
\multirow{2}*{\textbf{Scenarios}}&{\textbf{Time}}&{\textbf{Trace}}&\multirow{2}*{\textbf{Smartphones}}&{\textbf{Urban}}\\
&\textbf{Length}&\textbf{Distance}&&\textbf{Canyon}\\
\cline{1-5} 
&&&&\\[-0.8em]
{Training Data in}&\multirow{2}*{5.5 h}
&\multirow{2}*{650 km}&\multirow{2}*{Pixel 4}&\multirow{2}*{Light}\\
{Scenario I \& II}&&&&\\
\cline{1-5} 
&&&&\\[-0.8em]
{Testing Data in}&\multirow{2}*{1 h}
&\multirow{2}*{120 km}&\multirow{2}*{Pixel 4}&\multirow{2}*{Light}\\
{Scenario I}&&&&\\
\cline{1-5} 
&&&&\\[-0.8em]
{Testing Data in}&\multirow{2}*{0.5 h}
&\multirow{2}*{11 km}&\multirow{2}*{Pixel 4}&\multirow{2}*{Medium}\\
{Scenario II}&&&&\\
\cline{1-5} 
\hline
\end{tabular}
\label{tab_data}
\end{table}

\section{Experiments}
\subsection{Datasets}
To evaluate E2E-PrNet, we employ the open dataset of Android raw GNSS measurements for Google Smartphone Decimeter Challenge (GSDC) 2021 \cite{fu2020android}. After removing the degenerate data, we select twelve traces for training and three traces for inference, all of which are collected by Pixel 4 on highways or suburban areas. The ground truth of the smartphone locations is captured by a NovAtel SPAN system. As shown in Fig. \ref{fig:Scenarios}, we design two localization scenarios--fingerprinting and cross trace--to investigate our prospect that a model pre-trained in an area can be distributed to users in the exact area to improve GPS localization. Scenarios I and II utilize the same training data. In Scenario I, the testing data were collected along the same routes as the training data, but on different dates. Conversely, in Scenario II, we employ entirely different data collected from routes distinct from those of the training data to test our method. Further details on data splitting can be found in Table \ref{tab_data} and Appendix \ref{sec: data files}.

\subsection{Implementations}
\subsubsection{Implementations of E2E-PrNet}
We use Pytorch, d2l \cite{zhang2021dive}, and Theseus \cite{pineda2022theseus} libraries to implement our proposed E2E-PrNet. The neural network module in the end-to-end framework is implemented as a multilayer perceptron with 20 hidden layers and 40 neurons in each layer, according to \cite{weng2024prnet}. About the configuration of Theseus, we use a Gauss-Newton optimizer with a step size of 0.5 and 50 loop iterations. The optimizer uses a dense Cholesky solver for forward computation and backpropagtes gradients in the unrolling mode. To validate our strategy for dealing with missing receiver clock offset labels (RCOL), we also train an E2E-PrNet using location ground truth only. 

\subsubsection{Implementations of Baseline Methods}
We choose the WLS method as the baseline model-based method and implement it according to \cite{weng2023localization}. We also compare our proposed framework with set transformer, an open source SOTA end-to-end deep learning approach to improve positioning performance over Android raw GNSS measurements \cite{kanhere2022improving}. We set the key argument in set transformer--the magnitude of initialization ranges $\mu$--according to the 95\textsuperscript{th} percentile of the WLS-based localization errors. The weights of our trained set transformer are available at \url{https://github.com/ailocar/deep_gnss/data/weights/RouteR_e2e}.

\begin{table}[t]
\centering
\caption{Horizontal positioning scores of end-to-end solutions}
\begin{tabular}{c|cc}
\hline
\multirow{2}*{\textbf{Methods}}&\multicolumn{2}{c}{\textbf{Horizontal Score (meter)}$\downarrow$}\\
&\textbf{Scenario I}&\textbf{Scenario II}\\
\cline{1-3}
&&\\[-0.8em]
\textbf{WLS} & 16.390&20.666\\
&&\\[-0.8em]
\textbf{Set Transformer}& 9.699 ($\mu=15$m)&19.247 ($\mu=22$m)\\
&&\\[-0.8em]
\textbf{E2E-PrNet (No RCOL)} & {7.239}&{19.158}\\
&&\\[-0.8em]
\textbf{E2E-PrNet} & \textbf{6.777}&\textbf{18.520}\\
\hline
\end{tabular}
\label{tab1}
\end{table}

\begin{figure}[t]
\centerline{\includegraphics[width=0.4\textwidth]{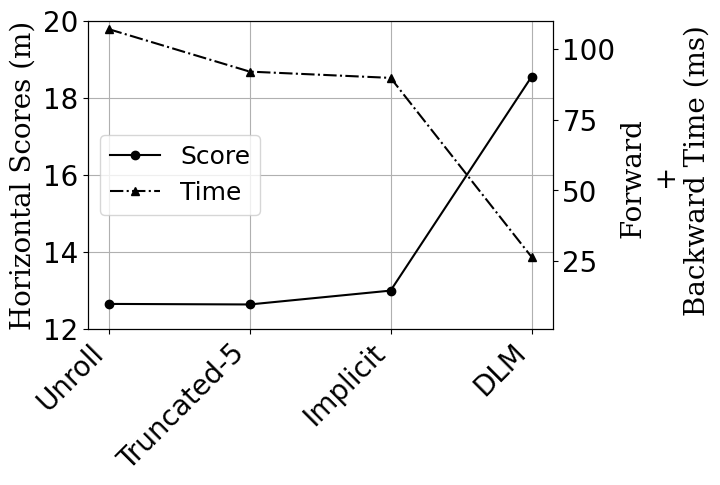}}
\caption{Backward mode analysis for configuring Theseus.}
\label{fig:backwardAnalysis}
\end{figure}

\begin{figure*}[t]
\centerline{\includegraphics[width=1\textwidth]{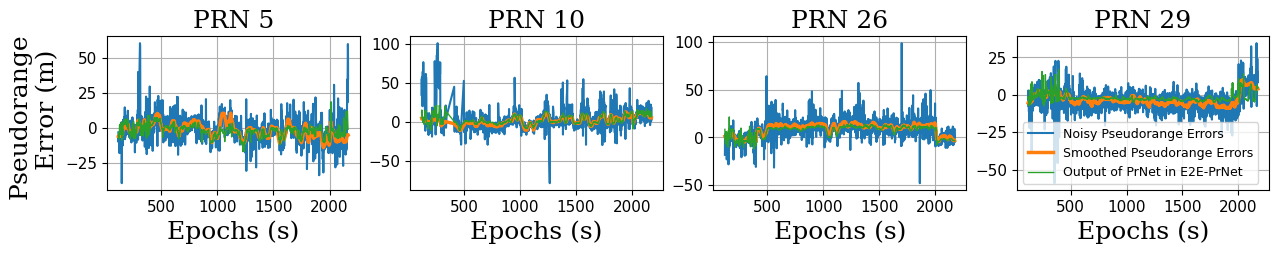}}
\caption{Output of the front-end PrNet in the E2E-PrNet framework.}
\label{fig:OutputPrNet}
\end{figure*}

\begin{figure*}[t]
\centerline{\includegraphics[width=1\textwidth]{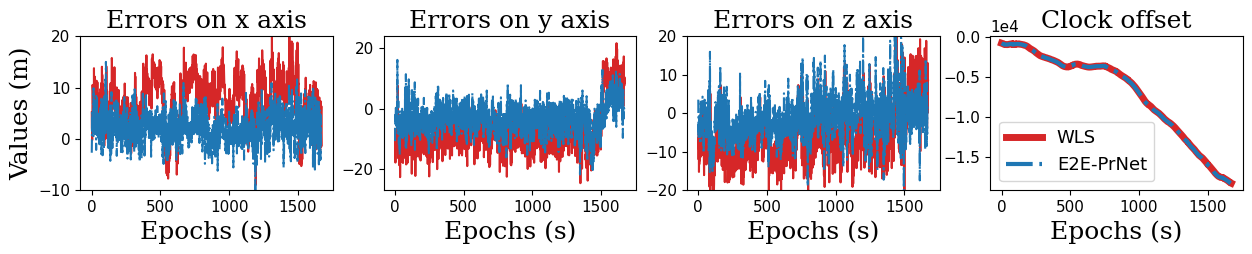}}
\caption{State estimations of WLS and E2E-PrNet.}
\label{fig:WLSvsE2ePrNet}
\end{figure*}

\subsection{Horizontal Errors}
The horizontal error is a critical indicator of GPS positioning performance in our daily applications. Google employs the horizontal errors calculated with Vincenty’s formulae as the official metric to compare localization solutions in GSDC. We show the horizontal errors of our proposed E2E-PrNet and the baseline methods in Fig. \ref{fig:RMSE}. Fig. \ref{fig: RMSE1} demonstrates that all three data-driven methods significantly reduce horizontal positioning errors in the fingerprinting scenario compared to the WLS algorithm. Among these methods, our proposed E2E-PrNet exhibits the best horizontal positioning performance. Regarding the cross-trace localization results--generalization results across different locations shown in Fig. \ref{fig: RMSE2}--despite a slight improvement, E2E-PrNet still outperforms its counterparts. For a clearer comparison, we present the empirical cumulative distribution function (ECDF) of their horizontal errors in Fig. \ref{fig:ECDF} and summarize in Table \ref{tab1} their horizontal scores, which are defined as the mean of the 50\textsuperscript{th} and 95\textsuperscript{th} percentile of horizontal errors. Among the end-to-end localization methods, our proposed E2E-PrNet obtains the best horizontal positioning performance in the two scenarios, with its ECDF curves generally being on the left side of those of other solutions. Quantitatively, E2E-PrNet achieves 59\% and 10\% improvement in horizontal scores compared to WLS. And its scores are also smaller than the set transformer by 30\% and 4\%. Compared with E2E-PrNet trained without RCOL, E2E-PrNet trained with the WLS-based estimation of receiver clock offsets obtains better horizontal positioning performance.

\subsection{Discussion on Backward Modes}\label{sec:Theseus}
Theseus designs four backward modes for training, including unrolling differentiation, truncated differentiation, implicit differentiation, and direct loss minimization (DLM). Their training performance varies across different applications \cite{pineda2022theseus}. Thus, we compare their training time and final horizontal localization scores during inference, as shown in Fig. \ref{fig:backwardAnalysis}. It indicates that the unrolling mode is the best regarding inference accuracy. Generally, the unrolling, truncated-5 (only the last five iterations are used), and implicit modes share similar horizontal scores. The truncated-5, implicit, and DLM modes have shorter training time than the basic unrolling mode because they have shorter backpropagation steps. DLM obtains the fastest training speed but the largest horizontal errors. Additionally, we notice that it is necessary to tune carefully the hyperparameters of DLM to keep its horizontal score small.

\begin{table}[t]
\centering
\caption{Comparison between E2E-PrNet and PrNet}
\begin{tabular}{c|cc}
\hline
\multirow{2}*{\textbf{Methods}}&\multicolumn{2}{c}{\textbf{Horizontal Score (meter)}$\downarrow$}\\
&\textbf{Scenario I}&\textbf{Scenario II}\\
\cline{1-3} 
&&\\[-0.8em]
\textbf{E2E-PrNet} & 6.777&18.520\\
&&\\[-0.8em]
\textbf{PrNet+Noisy Labels} & 6.922&18.434\\
&&\\[-0.8em]
\textbf{PrNet+Smoothed Labels} & 6.537&19.524\\
\hline
\end{tabular}
\label{tab2}
\end{table}

\subsection{Discussion on Explainability}
To verify whether the front-end PrNet in E2E-PrNet behaves as we expected, we record its output data during inference--the input to the downstream DNLS optimizer--and draw them together with the noisy and smoothed pseudorange errors in Fig. \ref{fig:OutputPrNet}. Here, we only show the results of four satellites; the other visible satellites share a similar phenomenon. While \eqref{eq:labelE2EPrNet} indicates the front-end PrNet should be trained under noisy pseudorange errors, Fig. \ref{fig:OutputPrNet} shows it approximately learns the smoothed pseudorange errors. This phenomenon can be explained by the observations that deep neural networks are robust to noise and can learn information from noisy training labels \cite{rolnick2017deep}. Then, the output of the front-end PrNet is approximately 
\begin{equation}\label{eq:realLabelE2ePrNet}
    \hat{\varepsilon}_k^{(n)}=\mu_k^{(n)}-\mathbf{h}_{t_k}^T\mathbf{M}_k.
\end{equation}

Substituting \eqref{eq:PrEq} and \eqref{eq:realLabelE2ePrNet} into \eqref{eq:residual} and letting \eqref{eq:residual} be zero yield
\begin{equation*}
    ||\hat{\mathbf{x}}_k-\mathbf{x}_k^{(n)}||+(\delta \hat{t}_{u_k}-\mathbf{h}_{t_k}^T\mathbf{M}_k)=\rho_{c_k}^{(n)}+\upsilon_k^{(n)}
\end{equation*}
where $\rho_{c_k}^{(n)}=\rho_{k}^{(n)}-{\varepsilon}_k^{(n)}$ is the clean pseudorange with its measurement error removed. According to \eqref{eq:exk}, the state estimation is
\begin{gather*}
\hat{x}_k=x_k+\mathbf{h}_{x_k}^T\mathbf{\Upsilon}_k,\hat{y}_k=y_k+\mathbf{h}_{y_k}^T\mathbf{\Upsilon}_k,\hat{z}_k=z_k+\mathbf{h}_{z_k}^T\mathbf{\Upsilon}_k,
\\
\delta \hat{t}_{u_k}=\delta t_{u_k}+\mathbf{h}_{t_k}^T\boldsymbol{\varepsilon}_k
\end{gather*}
where $\mathbf{h}_{x_k}^T$, $\mathbf{h}_{y_k}^T$, and $\mathbf{h}_{z_k}^T$ are the $1^{st}$-$3^{rd}$ row vectors of $\mathbf{H}_k$. Thus, the final output states should be approximately the WLS-based location estimates with biased errors removed and the WLS-based receiver clock offset estimates, which is verified by Fig. \ref{fig:WLSvsE2ePrNet}. It shows that the localization errors of E2E-PrNet on $x$, $y$, and $z$ axes fluctuate more closely around the zero level compared to the WLS-based solutions. More accurate and unbiased localization has been achieved through neural pseudorange correction. Furthermore, E2E-PrNet can still provide the basic receiver clock offset estimates as well as the WLS algorithm. E2E-PrNet behaves exactly as we expected.

We also compare the horizontal localization performance of E2E-PrNet with that of PrNets trained with noisy and smoothed labels, as displayed in Table \ref{tab2}, which shows their similar horizontal scores. Our preceding analysis claims that the front-end PrNet in E2E-PrNet is equivalently trained by noisy pseudorange errors \eqref{eq:labelE2EPrNet}. Consequently, the proposed E2E-PrNet performs as well as PrNet trained with noisy labels. Meanwhile, thanks to the robustness of deep neural networks to label noise \cite{li2020gradient}, both E2E-PrNet and PrNet trained with noisy labels share a similar performance to PrNet trained with smoothed labels. However, such equivalence among them might break down when the carrier-to-noise density ratio is extremely low--the intense pseudorange noise exists \cite{song2022learning}.

\section{Conclusion}
This paper explores the feasibility of training a neural network to correct GPS pseudoranges using the final task loss. To this end, we propose E2E-PrNet, an end-to-end GPS localization framework composed of a front-end PrNet and a back-end DNLS module. Our experiments on GSDC datasets showcase its superiority over the classical WLS and SOTA end-to-end methods. E2E-PrNet benefits from PrNet's superior ability to correct pseudoranges while mapping raw data directly to locations. 

Potential future work includes: 1) extending E2E-PrNet to other satellite constellations, such as BDS, GLONASS, and Galileo; 2) integrating carrier phase measurements into E2E-PrNet for more precise positioning \cite{liu2023performance}; 3) incorporating classical filtering algorithms and receiver dynamic models into E2E-PrNet for noise suppression \cite{wang2023neural}, and studying its feasibility in urban canyons with weak signal strength.

\appendices
\section{Training and testing data files}\label{sec: data files}
Table \ref{tab_training} and Table \ref{tab_testing} list the data filenames in the GSDC 2021 dataset, which we utilized for training and testing.

\begin{table}[ht]
\centering
\caption{Training data filenames}
\begin{tabular}{c|c}
\hline
{\textbf{Scenarios}}&{\textbf{Filenames}}\\
\cline{1-2} 
&\\[-0.8em]
\multirow{12}*{\textbf{Scenario I \& II}} & 2020-05-15-US-MTV-2\\
&\\[-0.8em]
& 2020-05-21-US-MTV-1\\
&\\[-0.8em]
& 2020-05-21-US-MTV-2\\
&\\[-0.8em]
& 2020-05-29-US-MTV-1\\
&\\[-0.8em]
& 2020-05-29-US-MTV-2\\
&\\[-0.8em]
& 2020-06-04-US-MTV-1\\
&\\[-0.8em]
& 2020-06-05-US-MTV-1\\
&\\[-0.8em]
& 2020-06-05-US-MTV-2\\
&\\[-0.8em]
& 2020-06-11-US-MTV-1\\
&\\[-0.8em]
& 2020-07-08-US-MTV-1\\
&\\[-0.8em]
& 2020-08-03-US-MTV-1\\
&\\[-0.8em]
& 2020-08-06-US-MTV-2\\
\hline
\end{tabular}
\label{tab_training}
\end{table}

\begin{table}[ht]
\centering
\caption{Testing data filenames}
\begin{tabular}{c|c}
\hline
{\textbf{Scenarios}}&{\textbf{Filenames}}\\
\cline{1-2} 
&\\[-0.8em]
\multirow{2}*{\textbf{Scenario I}} & 2020-05-14-US-MTV-1\\
&\\[-0.8em]
& 2020-09-04-US-SF-2\\
&\\[-0.8em]
\cline{1-2} 
&\\[-0.8em]
\textbf{Scenario II}&2021-04-28-US-MTV-1\\
\hline
\end{tabular}
\label{tab_testing}
\end{table}

\bibliographystyle{IEEEtran}
\bibliography{main}

\end{document}